\documentclass[10pt]{article}
\usepackage[a4paper,centering,hmargin=4cm,vmargin=3.5cm]{geometry}

\usepackage{graphicx}
\usepackage{dcolumn}
\usepackage{bm}
\usepackage[utf8]{inputenc}
\usepackage[english]{babel}
\usepackage[font=footnotesize,labelfont=bf]{caption}
\usepackage{amsmath,amssymb,amsfonts,latexsym,cancel, mathtools,wasysym,mathrsfs}
\usepackage{subcaption}
\usepackage{marvosym}
\usepackage{wrapfig}
\usepackage{float}
\usepackage{siunitx}
\PassOptionsToPackage{hyphens}{url}
\usepackage{hyperref}
\usepackage{authblk}
\usepackage{abstract}
\usepackage{todonotes} 
\usepackage{lipsum}
\usepackage{tipa}
\usepackage{verbatim}

\AtBeginDocument{}

\makeatletter
\def\@maketitle{%
  \newpage
  \null
  \vskip 0em%
  \begin{flushleft}
  \let \footnote \thanks
    {\Large\bfseries \@title \par}%
    \vskip 0.5em%
    {\quad\normalsize
      \lineskip 0.5em%
        \@author}%
        
    {\quad\quad\small \@date}%
  \end{flushleft}%
  \par
  \vskip 1.5em}
\makeatother

\makeatletter
\renewcommand\section{\@startsection {section}{1}{\z@}%
                                   {-3.5ex \@plus -1ex \@minus -.2ex}%
                                   {2.3ex \@plus.2ex}%
                                   {\normalfont\large\bfseries}}
\renewcommand\subsection{\@startsection{subsection}{2}{\z@}%
                                     {-3.25ex\@plus -1ex \@minus -.2ex}%
                                     {1.5ex \@plus .2ex}%
                                     {\normalfont\large\bfseries}}
\renewcommand\subsubsection{\@startsection{subsubsection}{3}{\z@}%
                                     {-3.25ex\@plus -1ex \@minus -.2ex}%
                                     {1.5ex \@plus .2ex}%
                                     {\normalfont\large\bfseries}}
\makeatother

\begin{document}
\title{Pronunciation recognition of English phonemes /\textipa{@}/, /æ/, /\textipa{A}:/ and /\textipa{2}/ using Formants and Mel Frequency Cepstral Coefficients}

\author{Keith Y. Patarroyo and Vladimir Vargas-Calderón\thanks{Physics Department. National University of Colombia. \newline Corresponding author e-mail: vvargasc@unal.edu.co}}
\date{}

\maketitle
\vspace{-1.1cm}
\begin{abstract}
The Vocal Joystick Vowel Corpus, by Washington University, was used to study monophthongs pronounced by native English speakers. The objective of this study was to quantitatively measure the extent at which speech recognition methods can distinguish between similar sounding vowels. In particular, the phonemes /\textipa{@}/, /æ/, /\textipa{A}:/ and /\textipa{2}/ were analysed. 748 sound files from the corpus were used and subjected to Linear Predictive Coding (LPC) to compute their formants, and to Mel Frequency Cepstral Coefficients (MFCC) algorithm, to compute the cepstral coefficients. A Decision Tree Classifier was used to build a predictive model that learnt the patterns of the two first formants measured in the data set, as well as the patterns of the 13 cepstral coefficients. An accuracy of 70\% was achieved using formants for the mentioned phonemes. For the MFCC analysis an accuracy of 52 \% was achieved and an accuracy of 71\% when /\textipa{@}/  was ignored. The results obtained show that the studied algorithms are far from mimicking the ability of distinguishing subtle differences in sounds like human hearing does.
\end{abstract}
\begin{center}
\line(1,0){250}
\end{center}


\section{Introduction}

Throughout computing history, scientists have developed a vast amount of theories and algorithms for speech recognition that are widely known (e.g. see the work by Lee (1988)). Many of them are motivated by using some of the principles of the human ear operation (Davis \& Mermelstein, 1980). In recent years, the blooming of high-speed processing computers has allowed us to start using deep machine learning to improve the efficacy of our speech recognizers (Baker et al., 2009). The techniques that are currently used for speech recognizers yield high prediction rates (Hinton et al., 2012). The most common and successful technique is the implementation of multi-layered neural networks (Zegers, 1998; Wellekens, 1998). The performance of such techniques seeded the question of implementing recognizers as objective evaluators of speech ability in humans.

This takes great relevance in the field of pronunciation teaching, not only because it provides the teachers (especially teachers who are non-native speakers of the taught language) a tool for assessing objectively the pronunciation of their students (Neri, Mich, Gerosa, \& Giuliani, 2008), and of themselves, but also because if a student has access to such a tool, he or she could learn pronunciation autonomously (Hinks, 2003). Unquestionably, the main goal in pronunciation teaching is to improve the ability of a language learner to use their vocal tract to produce sounds that are recognized as native sounds by native speakers of that language. Although this remains a challenge, particularly in the early stages of learning a language (Mirzaei, Gowhary, Azizifar, \& Esmaeili, 2015), it also provides a means by which students can improve the learning and retention of grammatical structures (Martin \& Jackson, 2016). Also, since different languages have different sound systems, it is normal for language learners to struggle learning them.

For instance, the phonemes /\textipa{@}/, /æ/, /\textipa{A}:/ and /\textipa{2}/ found in the English language are troublesome for many English learners because of their similarity. For example, Spanish and Italian speakers tend to mix these phonemes into a single one: /a/. There are cases of learners from other languages, such as Azerbaijani, in which this confusion has been studied (Ghaffarvand Mokari \& Werner, 2016).

This article aims to study the ability of two of the most historically important speech recognition tools to differentiate between these phonemes, i.e. how good tools would they be in evaluating the correct pronunciation of these phonemes. These tools are the formants (computed with Linear Predictive Coding), attractive for its simplicity in vowel pronunciation recognition, and the Mel Frequency Cepstral Coefficients, attractive for its computational speed in continuous speech recognition. Both tools were tested with the Vocal Joystick Vowel Corpus, from the Washington University.

\section{Theoretical Framework}
\subsection{Sound Formation}
The frequency range of the pressure waves in the air that compose the audible sounds by humans goes from 20\si{\hertz} to 20\si{\kilo\hertz} (Rosen \& Howell, 2011). The process by which sound is produced by humans is the following (for a complete understanding of sound production by humans see reference (O'Grady \& Dobrovolosky, 1997)). 

The diaphragm and intercostal contractions make the lungs generate a flow of air from the chest to the mouth. This air first passes through the larynx, wherein the vocal cords are. These are muscles that can be geometrically distributed in several ways, each of which is excited by the passing air creating modes of vibration or glottal states that result in sound. The pharynx, the oral cavity and the nasal cavity are filters and resonators of the aforementioned sounds. Also, the tongue and lips allow us to rapidly articulate and change the shape of the vocal cavity in order to filter some frequencies.

In this study, we are interested in vowels, which are voiced glottal states produced with little obstruction of the vocal tract. This means having the lips open and also the tongue without contact with the palate.

\subsection{Human Hearing Detection Principles}
It is also convenient to mention some of the principles that the human ear uses to detect sound signals. The human ear analyses pressure variations in the air into different frequencies. Roughly, the human ear does a Fourier transform of the pressure signal $P(t)$, where $t$ is the time, and transmits  $|P(\omega)|^2$, where $\omega$ is the frequency, to the brain (Sethna, 2006).

However, this initial model falls short for many analyses. Some of the reasons are (Lyons, n.d.): the tonal information is changing as a word or tune progresses; the difference between two closely spaced frequencies is hard to recognize for humans, especially at high frequencies; and humans hear loudness on a non-linear scale. All of these factors propose a significant challenge to mimic the human hearing system.

Davis and Mermelstein (1980) developed a method (Mel Frequency Cepstral Coefficients) to mimic this process, and has been the state of the art in the field of speech recognition since then.

\subsection{Formants and Linear Predictive Coding (LPC)}

The vocal tract can be thought of as a vibrating cavity, or resonator, whose geometry (morphology) is constantly changing in continuous speech. This change in the geometry allows the superposition of different modes of vibration, and thereby different sounds.

Experiments have shown that for English vowels there are some characteristic frequencies of vibration called formants (Hillenbrand, Getty, Wheeler, \& Clark, 1994; Hunter \& Kebede, 2012; Deterding, 2006), that correspond to maxima of vibration, i.e. where the acoustic energy is focused. The first formant is roughly located from 0 to 1\si{\kilo\hertz}, while the second formant is located from 1\si{\kilo\hertz} to 2\si{\kilo\hertz}, and so on (these frequency boundaries are not rigid, because there might be some second formants below 1\si{\kilo\hertz}, as seen in the references). In the studies mentioned only the two first formants are taken into account for characterising each vowel. However, there are studies like (Prica \& Ilić, 2010) in which 3 formants are used to characterise the Serbian vowels in order to improve accuracy when performing classification of vowel sounds.

Formants can be found by performing an acoustic power spectrum of a sound signal (or a spectrograph of the signal), and identifying the peaks in the spectrograph. This can be done by computing the envelope of the signal's frequency spectrum. Since the envelope contains information about the energy peaks, the formants can be determined. To predict the envelope of a signal $s[n(\Delta t)]$ sampled in discrete time steps $n=0,1,2,\ldots$, the Linear Predictive Coding method was introduced (Deng \& O'Shaughnessy, 2003). The goal of LPC is to predict $s[n]$ with a linear combination of $s[n-1], s[n-2], \ldots, s[n-M]$, where $M$ is the integer that sets the number of predicting signal samplings. 

\subsection{Mel Frequency Cepstral Coefficients (MFCC)}

This method aims to mimic the procedure performed by the human hearing system to decode a sound signal. It can be summarised in the steps shown in table \ref{tab:mfcc} (Lyons, n.d.).

\begin{table}[H]
\centering
\caption{Steps of the MFCC method compared to the sound analysis  of the human body.}
\label{tab:mfcc}
\begin{tabular}{ >{\centering\arraybackslash}m{0.25in}  >{\centering\arraybackslash}m{2in}  >{\centering\arraybackslash}m{2in} }
\hline
Step    &   Human Detection &   Mel Frequency Coefficients \\                      \hline
1.  &   The hearing sense is sensible to the tonal information change as a word or tune progresses. & Frame the signal into short frames  to account that on short time scales the audio signal does not change much.\\

2.   & Different frequencies are detected by  vibrations at different spots of the cochlea depending on the frequency of the incoming sounds.             & For each frame calculate the periodogram estimate of the power spectrum.   \\ 
3.   & The cochlea cannot discern the difference between two closely spaced frequencies. & Apply the Mel filterbank to the power spectra, sum the energy in each filter. \\ 
4.   & Loudness is detected on a non-linear scale, where differences in small frequencies are  more considerable. & Take the logarithm of all filterbank energies.  \\ 
5.   &  --------------------------------- & Take the DCT (Discrete Cosine Transform) of the log filterbank energies.\\ 
6.   & Identifies the linguistic content and  discards all the other stuff (background noise, emotion) & Keep DCT coefficients 2-13, discard the rest.\\
\end{tabular}
\end{table}

\subsection{Decision Tree Classifiers (DTC)}

Classifiers are computer tools that learn patterns from labelled data, so that when a new sample is presented to the computer, it will classify the sample into one of the different labels. In other words, consider a data set $D = \{(d_i,l_i)\}$, where $d_i$ is a vector of $N$ components (features). The vector $d_i$ is also called a sample. $l_i$ is a label. The classifier is a function $f(d): F\to L$, where $F$ is the vector space of the samples and $L$ the space of the labels.

To see how DTCs work, consider the following example. Suppose that a data set consists of samples in one dimension: each sample is the salary of every single person in Colombia. The labels for this data set are the social stratum. We might have people that earn a lot of money, but they live in stratum 2, or vice versa. However, these cases are strange and in general there is a structure that allows predicting the label (social stratum) from the samples. The DTC will try to learn the salary ranges for each single social stratum, so that when a new sample is presented to the model, it will classify it in a specific label. The boundaries of the salary ranges are random at first, but as the learning process advances, the boundaries become more accurate and clear.

To test how well the DTC is able to predict labels, a data set $D'$ called the testing data is normally built. $D'$ consists of $M$ samples with known labels. Only the samples, and not the labels, are presented to the DTC. The percentage of correctly predicted labels will tell the accuracy or effectiveness of the DTC. For a more thorough explanation, refer to (Tan, Steinbach \& Kumar, 2005).

\section{Methodology}

\begin{figure}[H]
    \centering
    \includegraphics[width = 0.75\textwidth]{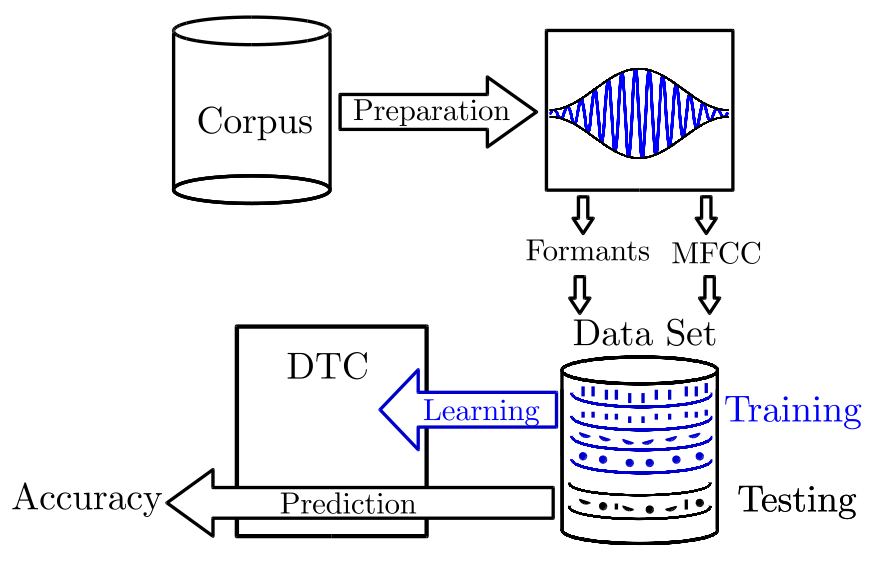}
    \caption{Diagram of the method used to measure the performance of formants and MFCC in pronunciation prediction.}
    \label{fig:pipeline}
\end{figure}

The pipeline of this study (see Figure~\ref{fig:pipeline}) was to extract and prepare audio files from the Vocal Joystick Vowel Corpus. Then, formants and cepstral coefficients were computed for each file. The set of files with their corresponding formants and cepstral coefficients was split into training and testing data. A DTC was trained with the training data, and tested with the testing data, yielding an accuracy percentage. The characteristics of each step are explained in this section.

\subsection{Corpus and Audio Processing}

In this study we used the Vocal Joystick Vowel Corpus (Bilmes, Wright, Xiao, Malkin Kilanski, 2006). This project selected a group of nine monophthongs and 12 vowel- to-vowel transitions. Sounds with different duration, amplitude and intonation were recorded for each vowel:
\begin{itemize}
    \item Duration: short (1 second), long (2 seconds) and nudge (very short repetitions of the same vowel).
    \item Amplitude: quiet, normal, loud, quiet to loud and loud to quiet.
    \item Intonation: level, rising and falling.
\end{itemize}

From this corpus, we considered all the sound files available in the corpus corresponding to the phonemes /\textipa{@}/, /æ/, /\textipa{A}:/ and /\textipa{2}/ whose duration was either short or long, whose amplitude was quite, normal or loud, and whose intonation was in a single level. A total of 784 files satisfied these conditions.

\subsection{Data Preparation}

\begin{figure}[H]
    \centering
    \includegraphics[width=\textwidth]{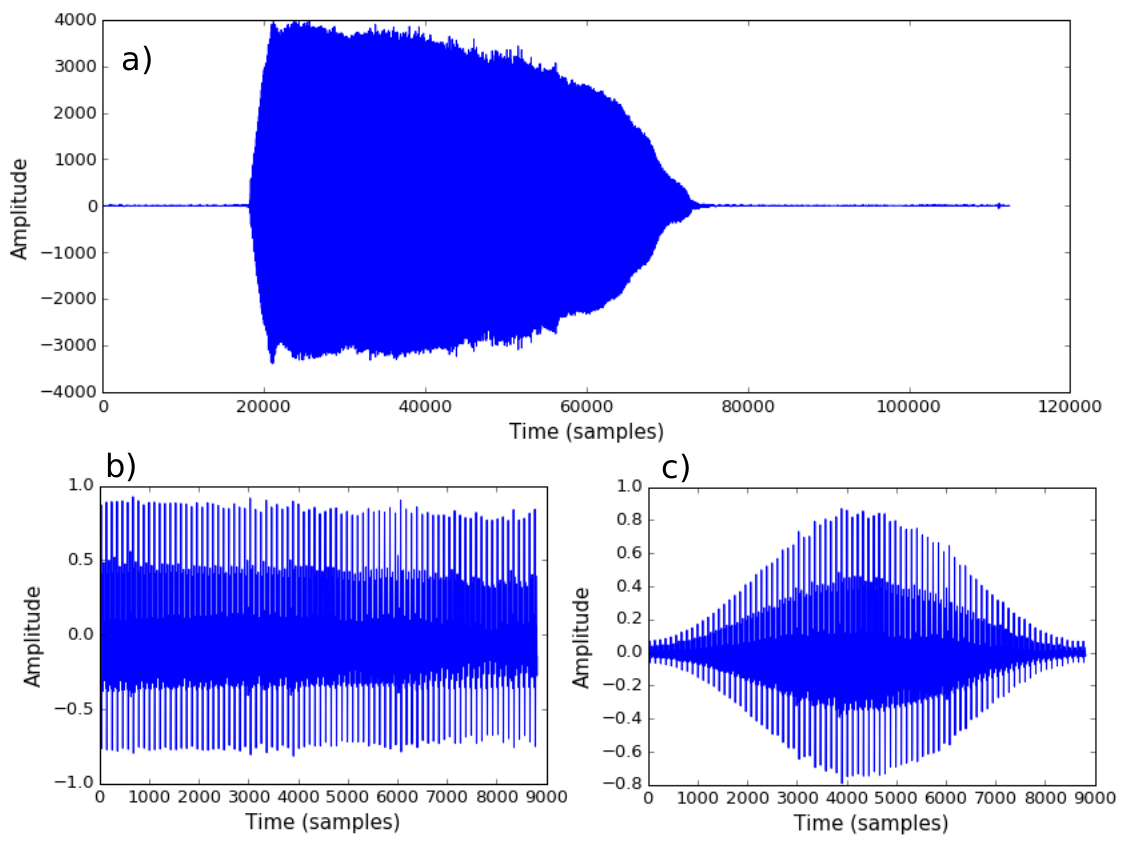}
    \caption{Preparation example of a sound signal corresponding to the /\textipa{2}/ phoneme. a) Raw signal. b) Silence deletion, amplitude normalisation and selection of a 40ms interval. c) Application of a Hamming window.}
    \label{fig:preparation}
\end{figure}

An example of a raw data file from the corpus is shown in Figure~\ref{fig:preparation}a. It can be seen that there are silent parts and the amplitude is not normalised. Also, even though we selected a single level of intonation, it is clear that the amplitude of the sound decreases as time progresses. Therefore, for each file we deleted the silent part. Then, a section of duration 40ms was selected so that the signal does not change considerably and is approximately periodic (Figure~\ref{fig:preparation}b). Finally, a Hamming window was applied to the resulting sound signal with the purpose of smoothing the Fourier Transform used in the MFCC extraction (Figure~\ref{fig:preparation}c).

\subsection{Evaluation of Formants and MFCC Performance with a DTC}

Then, the LPC Python implementation by Danilo Bellini was used to compute the formants (Bellini, 2016). To do the calculation of the MFCC, the Python implementation by James Lyons (Lyons, 2016) was used. Finally, the set of computed formants and the set of computed cepstral coefficients were partitioned into two groups each. One of the groups contained two thirds of the data, and the other one contained a third. The former was the training data, and the later was the testing data. To exemplify this procedure, denote $S=\{(s, v)\}$ as the set of the sound signal files, each of which is labelled by a vowel $v$. Let $F(s)$ be a function that computes the formants $F$ of a sound signal $s$. Let  $X=\{(F(s), v) \,\forall s\in S\}$ be the data set of formants, labelled by their respective vowel. Let $X_{tr}\subset X$ be the set of training data and $X_{te}\subset X$ be the set of testing data. The DTC learns the ranges corresponding to each vowel in the formants space by creating a function $DTC(F(s))$ that takes the values of the labels, i.e. the phonemes /\textipa{@}/, /æ/, /\textipa{A}:/ and /\textipa{2}/. For instance, if after the learning process is finished, a sample $F(s)$, labelled by the phoneme /\textipa{@}/, is presented to the DTC, and $DTC(F(s)) =$ /æ/, then the DTC failed to predict the correct phoneme. If instead $DTC(F(s)) =$ /\textipa{@}/ then the DTC was successful in the prediction. The accuracy is thus defined as the percentage of correctly predicted phonemes from the testing data. We used the accuracy to test the quality of prediction by the DTC using formants and Mel coefficients separately. This gives a measure of how well could formants and cepstral coefficients help in the objective evaluation of pronunciation.

\section{Results, Analysis and Discussion}

\subsection{Formants}

\begin{figure}[H]
    \centering
    \includegraphics[width=0.8\textwidth]{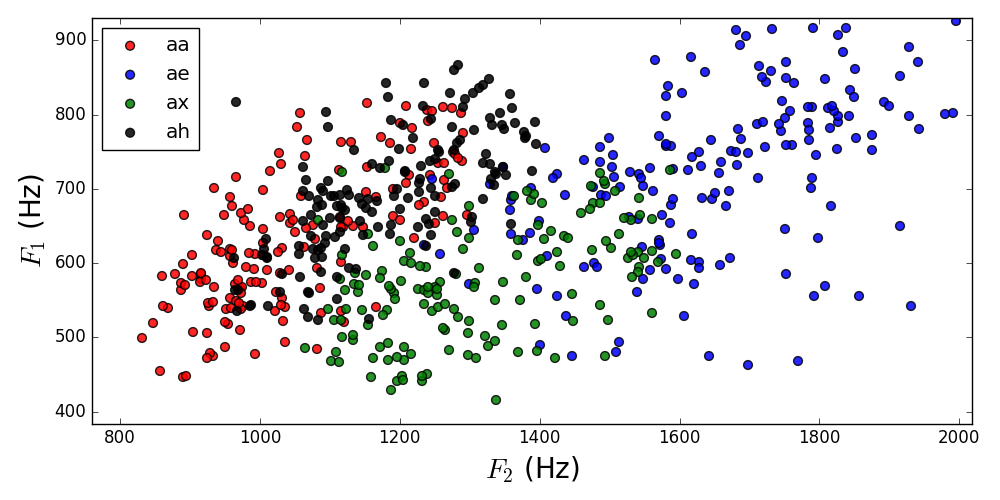}
    \caption{Frequencies in Hertz of the first two formants for the studied phonemes. aa is /\textipa{A}:/, ax is /\textipa{@}/, ae is /æ/ and ah is /\textipa{2}/.}
    \label{fig:formants}
\end{figure}

From the 784 sound files, 142 were discarded for this analysis and for the one corresponding to the MFCC. The reason to discard these files was that some recorded sounds contained parts in which the signal was not periodic due to a trembling voice from the speaker. This could cause the LPC and MFCC algorithms to be affected. The criterion used to discard the data is now explained. Denote the recordings or sound signals that belong to vowel $v$ as $(s,v)$. Also, denote the mean value of the $i$-th formant ($i=1,2$) and the standard deviation of the $i$-th formant, for a vowel $v$, as $(\mu_i,v), (\sigma_i, v)$, respectively. The remaining 642 sound signals satisfied the condition
\begin{align}
    |F_i(s)-\mu_i|<1.5 \sigma_i,\label{eq:criterion_formants}
\end{align}
for each vowel $v$, where $F_i$ refers to the $i$-th formant. The computed formants for these sound signals are shown in figure \ref{fig:formants}.

The DTC trained with the formants yielded a 70\% accuracy. Although this success rate of the predictive model is a very low rate, it can be considered as good taking into account that only two features were used to predict the data: the first two formants. This means that with little information about four very similar sounding phonemes, we were able to predict, with an accuracy of 70\%, the vowel of a sound taken at random from the recordings obtained from the corpus.

Furthermore, Figure~\ref{fig:formants} shows a non-Gaussian distribution in either formant for every phoneme. This is supported by observations in which it was seen that formants do not only depend on the geometry of the vocal tract of each person, but also are statistically dependent on the age, as shown in (Hawkings \& Midgley, 2005); as well as on gender, as shown in (Hillenbrand, Getty, Clark, \& Wheeler, 1995). The formants measured are shown in table \ref{tab:formants} and compared with measurements from other studies.

\begin{table}[H]
\centering
\caption{Formants measurement in Hertz. Our measurements are compared with Hawkings \textit{et. al.}, Hunter \textit{et. al.}, Deterding.\label{tab:formants}}
\begin{tabular}{c|cc|cc|cc|cc}
\hline
Study & \multicolumn{2}{|c|}{Our} &\multicolumn{2}{|c|}{Hawkings \textit{et. al.}} & \multicolumn{2}{|c|}{Hunter \textit{et. al.}} & \multicolumn{2}{|c}{Deterding} \\[4pt]
\hline
Phoneme & $F_1$ & $F_2$ & $F_1$ & $F_2$ & $F_1$ & $F_2$ & $F_1$ & $F_2$\\
\hline
/\textipa{@}/ & 631 & 1049 & 496 & 833 & 643 & 1019 & 625 & 973\\
/æ/ & 720 & 1644 & 696 & 1574 & 667 & 1565 & 748 & 1360\\
/\textipa{A}:/ & 573 & 1311 & 608 & 1062 & 680 & 1193 & 757 & 1211\\
/\textipa{2}/ & 693 & 1182 & 629 & 1160 & 661 & 1296 &724 &1282\\
\hline
\end{tabular}
\end{table}

Studies shown in Table~\ref{tab:formants} show that formants vary a lot with age, gender and geography. Therefore, we see that identifying a phoneme by its formants is not trivial, but as we showed, can be done with an estimated accuracy of 70\%. Despite the low accuracy, formants keep being used for vowel pronunciation research, as in (Rasilo \& Räsänen, 2017), in which the process of infants learning their native language was simulated through a Leaning Virtual Infant that received sounds from its caregivers. At first the Learning Virtual Infant babbled randomly, but as the learning process advanced, it began to compute formants to identify similar sounding vowels and to improve its pronunciation.

\subsection{MFCC}

\begin{figure}[H]
    \centering
    \includegraphics[width = \textwidth]{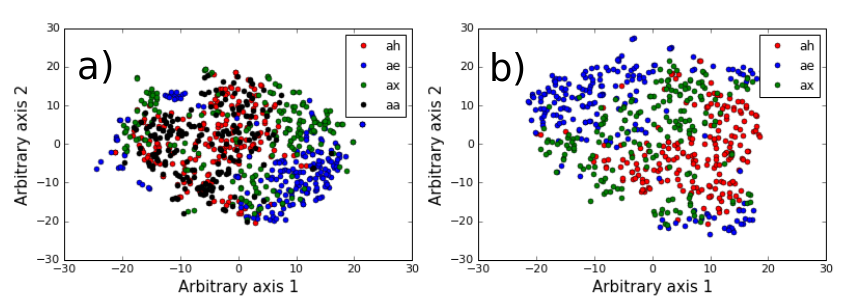}
    \caption{PCA of the MFCC. a) shows the four studied phonemes and b) shows only /\textipa{@}/, /æ/ and /\textipa{2}/.}
    \label{fig:mfcc}
\end{figure}

13 MFCC were obtained for every sound signal. In order to visualise the MFCC of every sound signal, a dimension reduction from 13 to 2 (i.e. from the MFCC space to a plane) was performed with Principal Component Analysis (PCA). This method projects the data from the high-dimensional space onto a plane that preserves the maximum possible variance of the data. The distribution of the MFCC for the different phonemes on the plane computed with PCA is shown in Figure~\ref{fig:mfcc}.

As it can be seen from Figure~\ref{fig:mfcc}a, the MFCC of the phonemes seem mixed. In particular, note that the points corresponding to the phoneme /\textipa{A}:/ are mixed with the ones corresponding to the phoneme /\textipa{2}/. This can also be seen in the formants figure. Ignoring the /\textipa{A}:/ points produces Figure~\ref{fig:mfcc}b, in which the clusters of data can be identified.

Partitioning the data shown in Figure~\ref{fig:mfcc}a in training and testing data, an accuracy of 56\% was accomplished with the DTC. While if the data shown in Figure~\ref{fig:mfcc}b was used, an accuracy of 71\% was measured. These very low accuracies show that the MFCC method lacks sensibility to subtle differences between the studied sound signals. 

\begin{figure}[H]
    \centering
    \includegraphics[width=\textwidth]{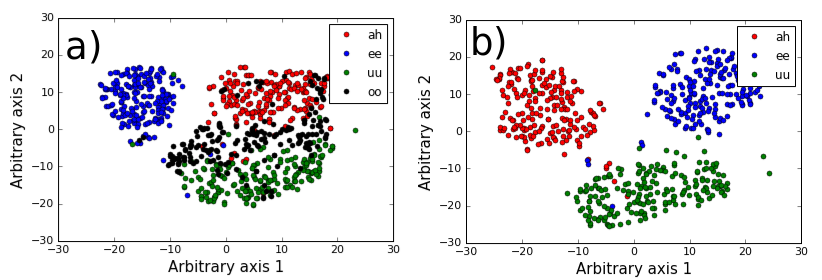}
    \caption{PCA of the MFCC. a) shows the phonemes /\textipa{2}/, /e/, /u/ and /o/. b) shows only /\textipa{2}/, /e/ and /u/. ee is /e/, uu is /u/ and oo is /o/.}
    \label{fig:allvowels}
\end{figure}

But, why are the MFCC so good at speech recognition then? The first thing to mention is that the MFCC are normally used to train models that are more complex than a DTC, like neural networks. The goal in speech recognition is to understand words, instead of single sounds. To check the accuracy of the MFCC at recognising different sounding sounds, we proceeded to analyse the phonemes /\textipa{2}/, /e/, /u/ and /o/. The PCA results are shown in Figure~\ref{fig:allvowels}.

Using the data shown in Figure~\ref{fig:allvowels}a, the accuracy was measured in 73\%. Clearly the MFCC corresponding to the phoneme /o/ were mixed with the ones corresponding to the phoneme /u/. To avoid this noise, if /o/ is removed, an accuracy of 88\% was reached. This shows that MFCC are a reasonable method to recognise different phonemes.

\section{Conclusions and Perspectives}

The Vocal Joystick Corpus was used to build a data set of sounds corresponding to the phonemes /\textipa{@}/, /æ/, /\textipa{A}:/ and /\textipa{2}/. From each of the sound signals, both formants and MFCC were computed. These were used to train a DTC that acted as a classifier. Using formants, an accuracy of 70\% was achieved. Using MFCC, an acuraccy of 52\% was measured. MFCC's performance was much better when the DTC was trained with the coefficients computed from the phonemes /\textipa{2}/, /e/ and /u/, reaching 88\%. However, these speech recognition tools fail to correctly predict the phonemes of similar sounding signals. 

As a response to this deficiency, computer and linguistics scientists work to build computational tools that learn subtle differences between sounds in order to assess a person’s pronunciation correctly. For instance, a research in the Chinese language showed results that improved significantly the previous recognition models in detecting mispronunciation of phonemes (Wei, Hu, Hu, \& Wang, 2009). However, the advance of technology to support pronunciation learning must be accompanied by designed learning programs that help students to learn autonomously. More people need to be encouraged to work in this interdisciplinary field that is pioneering and improving language teaching.

\end{document}